\documentclass[11pt]{article}

\usepackage{graphicx}

\usepackage{standalone}

\usepackage{tikz}
\usetikzlibrary{calc}
\usetikzlibrary{positioning}
\usetikzlibrary{shapes.geometric}
\usetikzlibrary{decorations.pathreplacing}
\newcommand*{\vcenteredhbox}[1]{\begingroup
\setbox0=\hbox{#1}\parbox{\wd0}{\box0}\endgroup}
\definecolor{matplotblue}{HTML}{1F77B4}

\usepackage{titling}
\usepackage[a4paper,margin=1in]{geometry}

\begin{document}

\title{One-Shot Identification with different neural network approaches}

\author{
Janis Mohr\\
Interdisciplinary Institute for Applied Artificial Intelligence and Data Science Ruhr,\\
Bochum University of Applied Sciences, Germany\\
ORCID: {0000-0001-6450-074X}
\and
J\"org Frochte\\
Interdisciplinary Institute for Applied Artificial Intelligence and Data Science Ruhr,\\
Bochum University of Applied Sciences, Germany\\
ORCID: {0000-0002-5908-5649}
}

\date{}

\maketitle              

\begin{abstract}
Convolutional neural networks (CNNs) have been widely used in the computer vision community, significantly improving the state-of-the-art. But learning good features often is computationally expensive in machine learning settings and is especially difficult when there is a lack of data.
One-shot learning is one such area where only limited data is available. In one-shot learning, predictions have to be made after seeing only one example from one class, which requires special techniques.
In this paper we explore different approaches to one-shot identification tasks in different domains including an industrial application and face recognition. We use a special technique with stacked images and use siamese capsule networks.
It is encouraging to see that the approach using capsule architecture achieves strong results and exceeds other techniques on a wide range of datasets from industrial application to face recognition benchmarks while being easy to use and optimise.

\end{abstract}

\section{Introduction}
\label{sec:introduction}
Mammals have the ability to recognise patterns and acquire new skills. In particular, humans can decide, whether two objects are identical or belong to the same category.
For example, while humans are capable of perceiving that two different breeds of cats belong to the same category, they can also easily distinguish between a cat and a dog. This skill, more than any other, serves as an essential survival skill, particularly as it relates to identifying food and avoiding dangerous entities.
Humans have the ability to quickly and accurately categorise species, even though they may have never seen an individual from the species before — all based on their memories and experience \cite{Kuhl.2020}. Little or even no additional data is needed.

Popular machine learning tasks and settings, on the other hand, tend to use very large amounts of data for training but this does not necessarily prevent them from failing when they are shown new and unfamiliar data. Generalisation is a problem of its own.

A challenge of particular interest is the training of machine learning models with limited training data. One-shot learning refers to the idea that an image of a class is only ever seen once. One-shot identification is a special case. Determining whether or not two items belong to the same class is the task at hand. For instance, after being shown images of an object through a series of various manufacturing processes, a neural network would be able to discern the object.

Computer vision methods can be used for a wide variety of tasks. Not only image recognition but also quality assurance. A special use-case in industrial vision is the inspection of produced objects. For industrial tasks, it often is complicated to acquire large amounts of training data because it can be difficult to mount cameras or other sensors during ongoing production.
Furthermore convolutional neural networks are used for face recognition tasks. With facial recognition, special attention must always be paid to data protection concerns, which also makes it particularly difficult to obtain larger data sets. This makes these two areas in particular common tasks in the field of one-shot learning, which is under special focus in research and of significant relevance in industrial application.

\subsection{Related Work}

Overall, research into one-shot learning algorithms has received limited attention by the machine learning community. Despite this, there are a few key papers on this task.
Initial work on this topic dates back to the years 2003 and 2006\cite{FeiFei.2003,FeiFei.2006}. The authors developed a variational Bayesian framework for one-shot image classification using the premise that previously learned classes can be leveraged to help forecast future ones when very few examples are available from a given class. 
In a series of papers, the problem of one-shot learning was approached from the viewpoint of cognitive science, proposing a method called Hierarchical Bayesian Program Learning which focuses on character recognition\cite{Lake2011OneSL,Lake2012ConceptLA,Lake2013OneshotLB}. This method aims at decomposing the image into small pieces and determining a structural explanation for the observed pixels.
Some researchers have considered transfer learning approaches for recognising new words from unknown speakers\cite{Lake2014OneshotLO} and path planning algorithms \cite{Wu2012OneSL}.\\

Although the idea of siamese networks has been around for quite some time \cite{Bromley.1993}, it has only recently been used for one-shot learning \cite{Chicco.2020}. They have mostly been used for robust face recognition and verification as \cite{Chopra.2005,Taigman.2014} have shown.
\cite{Koch.2015} use siamese neural networks for one-shot image recognition in which they use two identical neural networks that are trained and then fed with two different images. The task is to determine whether those two images belong to the same class. \cite{Deshpande.2020} especially used siamese networks to detect defects in steel surfaces during manufacturing somewhat similar to the industrial application used as a benchmark in this paper.\\

Scientific work focused on a mathematical representation of the human visual cortex led to \cite{GeoffreyE.Hinton.1981} identifies that some of the processes in the human brain related to vision have similarities to the concept of inverse computer graphics. In inverse computer graphics, an image is analysed and the task is to locate entities in this image in such a way that complex objects are broken down into several simple geometric shapes \cite{GeoffreyE.Hinton.1981b}.
Currently used CNNs do not operate in a way that is consistent with this principle \cite{Tielenman.2014}. Building on initial experiments\cite{GeoffreyE.Hinton.2011}, so-called capsule networks and dynamic routing in which neurons are grouped into capsules and vectors are shared between layers were proposed \cite{DynamicRouting}.
Some additional work was done in different domains to study the performance and possibilities of capsule networks. Rajasegaran et al.\cite{Rajasegaran.2019,Xi.2017} analysed performance on key-benchmark datasets like Fashion-MNIST and CIFAR-10. Several papers work out that capsule networks surpass baseline CNNs in image classification tasks across several domains \cite{abeysinghe2021capsule,Afshar.2018,gagana2018activation,Iesmantas.2018,Kumar.2018,Mobiny.2018,Renkens.2018}.
\cite{Li.2019,Mohr.2021.capsules,Quetscher.2022,Renzulli.2022,Atefeh.2018} focused on the qualities regarding explainability which are based on the proximity to the functioning of human vision.\\

Extensive work has been done to identify objects by their structure. Minderer et al.\cite{Minderer2019} uses video clips to extract the structure of objects. Using and creating videos is often not possible due to constraints regarding to privacy or because of costs.
Abdel et al. \cite{Abdel.2003} studied various computer vision algorithms, including fast Haar transform and Canny edge detector for the detection of cracks and other kinds of damage in concrete. This study was enhanced by Cha et al. \cite{Cha.2017} which presented an approach to recognise defects on concrete surfaces. Instead of classic computer vision algorithms, they used a deep convolutional neural network and furthermore used a rule-based approach as a baseline. They summarise that it is possible to outperform classic machine vision techniques with neural networks and that these approaches can be very robust in situations where different lighting can occur.
Benhimane et al. \cite{Benhimane2008} uses a template management algorithm to track 3D objects in real-time. Brusey et al. and Mahieu et al. \cite{Brusey2009,Mahieu2019} propose to use RFID chips and vision systems to identify objects during manufacturing.\\

Mohr et al. \cite{Mohr21} proposed the approach for merged images used in this paper and also have an in-depth explanation for the industrial application. This paper extends Mohrs' work by looking at another and additional approach, the siamese capsule networks, and by conducting experiments on two other datasets from different domains.

Our contributions in this paper are summarised as follows: (1) We analyse approaches for identifying objects which have only been seen once with capsule networks. (2) We focus on techniques that are economical with data. (3) We show the versatility and high accuracy of the presented approaches and compare several approaches in different domains.

\section{Approach}

In this section, the three approaches used in this paper including the siamese capsule network are described and explained in detail.

\subsection{Classic Convolutional Neural Network with Merged Images}

Although deep learning has shown great promise for image recognition tasks, it comes with the cost of large data requirements. Since it is not always possible to gather large amounts of data due to e.g. information privacy, the possible applications for machine learning are still limited. To address this issue, there has been recent interest in the research community to develop neural networks, that can effectively learn from a small amount of data.

\begin{figure}
\begin{center}
\input{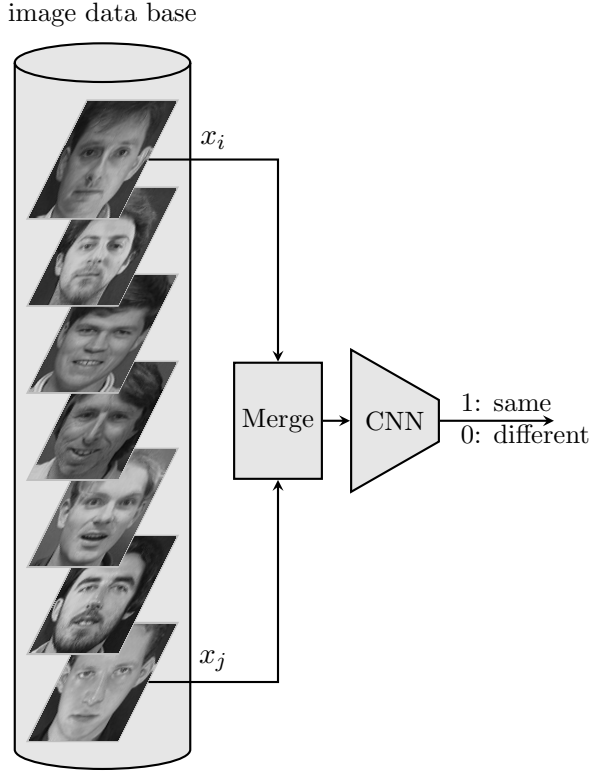}
\caption{General approach to one-shot identification using merged images. Two images are merged and then put into a CNN to classify whether they contain the same or different objects. The images on the left are a random selection from the AT\&T dataset used in Section~\ref{sec:Experiments}.
}
\label{fig:approach}
\end{center}
\end{figure}

One methodology to decrease the quantity of training data required while keeping reliable predictions is one-shot learning. One-shot image recognition's core tenet is that a neural network should be able to determine whether or not examples belong to the same class given a single sample of an image out of that class. Throughout training, the network determines which features of the input image pair deviate from one another. Every convolutional neural network may be employed with this method of one-shot identification, depending on the application. Once trained, the model ought to be able to recognise objects despite variations in colour or material properties. The network's output is [same, different]. {\itshape Same} denotes the presence of the same object in both photos, whereas {\itshape different} denotes the presence of different objects. The solution outlined by \cite{Mohr21} seeks to act in the unique situation when only very little training data is provided and comparable images are only seen once.

Figure \ref{fig:approach} shows the proposed approach with a data set. A classic convolutional neural network is trained on a data set that consists of images of human faces with different expressions and lighting. The task is to compare two images, which have never been seen before, and to decide if they show the same person or not. The network is trained with merged and labeled images, which show the same or different humans. The meaning of {\itshape merged} in practice in this method is explained in section \ref{sec:MergingTwoImagesInPreprocessing}. 
Therefore, the neural network learns to discriminate between different objects and can generalise this information to compare images of objects it has never seen before.

The hyperparameter values used for training the network are provided in table \ref{tab:param}. Four convolutional layers with 32, 32, 64, and 64 filters are used, followed by a max pooling \cite{Scherer.2010}. A kernel size of 3×3 is used for convolutions with a stride of 1. The ReLU activation function is used on the output feature maps of each layer. The convolutional layers are followed by two fully connected layers of size 128 and 2 respectively.

\begin{table}[!htbp]
\caption{Hyperparameter values used for training the convolutional neural network.\cite{Mohr21}}\label{tab:param}
\begin{center}
\begin{tabular}{|l|l|}
\hline
Parameter &  Value\\
\hline
Batch Size & 32\\
Number of Epochs & 20\\
Learning Rate & 1e-4\\
Optimiser & RMSprop \cite{Graves.2013}\\
\hline
\end{tabular}
\end{center}
\end{table}

\subsubsection{Handover of Two Images}
A basic CNN uses one picture at a time as input \cite{LeCun1998}. Finding a procedure that can take two inputs and compare two images is essential for the comparison of images. This results in two possible alternatives.
Two images are merged, and the combined picture is supplied into a CNN for training. The alternative method is based on utilizing two convolution and pooling sequences. The convolutional neural network receives two images as input. Following flattening, the two sequences—each supplied with a distinct image—are concatenated in a particular layer.

\begin{figure}[htb]
\vcenteredhbox{\begin{tikzpicture}[img box/.style={draw, xslant=0.5, minimum width=1.9cm, minimum height=2.5em}]
  \node [img box]                     (img 1) {};
  \node [img box, right=0pt of img 1] (img 2) {};
  \node [font=\small] at (img 1) {image 1};
  \node [font=\small] at (img 2) {image 2};
\end{tikzpicture}}
\hfill
\vcenteredhbox{\begin{tikzpicture}[img box/.style={draw, xslant=0.5, minimum width=1.9cm, minimum height=2.5em}]
  \node [img box]                     (img 1) {};
  \node [img box, below=0pt of img 1] (img 2) {};
  \node [font=\small] at (img 1) {image 1};
  \node [font=\small] at (img 2) {image 2};
\end{tikzpicture}}
\caption{Merging two images into a bigger image by joining them horizontally or vertically.\cite{Mohr21}} \label{fig:merging1:howto}
\end{figure}
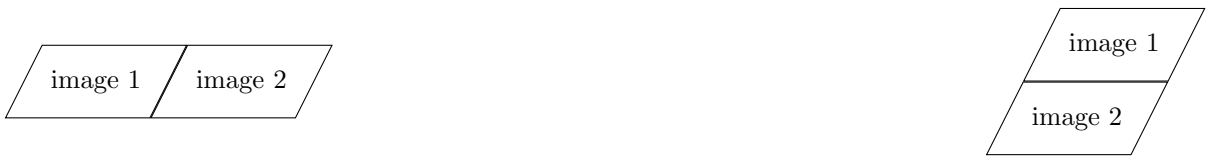

\label{sec:MergingTwoImagesInPreprocessing}
Several ways of merging the images are possible. One option is, that the images could be joined horizontally or vertically, as shown in figure \ref{fig:merging1:howto}. If the original dimensions are $n$ and $m$ the result is an $2n \times m$ or $n \times 2m$ image. 

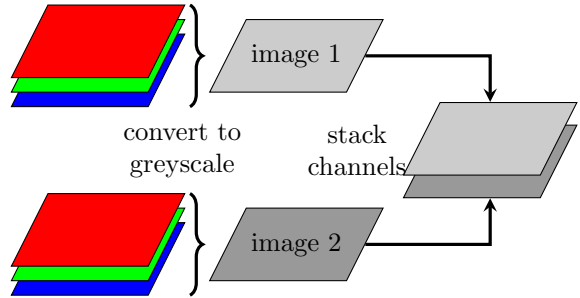
\begin{figure}[htb]
\begin{center}
\begin{tikzpicture}[img box/.style={draw, xslant=0.5, minimum width=1.8cm, minimum height=2.5em}]
  \begin{scope}[yshift=-1.25cm]
    \draw [fill=blue]  (0, 0)     -- ++(right:1.8) -- ++(1.25em, 2.5em) -- ++(left:1.8) -- cycle;
    \draw [fill=green] (0, 0.5em) -- ++(right:1.8) -- ++(1.25em, 2.5em) -- ++(left:1.8) -- cycle;
    \draw [fill=red]   (0, 1em)   -- ++(right:1.8) -- ++(1.25em, 2.5em) -- ++(left:1.8) -- cycle;
    
    \draw [decorate, decoration={brace, amplitude=1ex, mirror}, very thick] (2.35, 0) -- ++(0, 3.5em);
    
    \node [img box, fill=black!40] at (3.75, 1.75em) (img 2) {};
    \node [img box, fill=black!40] at (6.3, 1.75em+1.1cm) (img 2b) {};
    \node [font=\small] at (img 2) {image 2};
  \end{scope}

  \begin{scope}[yshift=1.25cm]
    \draw [fill=blue]  (0, 0)     -- ++(right:1.8) -- ++(1.25em, 2.5em) -- ++(left:1.8) -- cycle;
    \draw [fill=green] (0, 0.5em) -- ++(right:1.8) -- ++(1.25em, 2.5em) -- ++(left:1.8) -- cycle;
    \draw [fill=red]   (0, 1em)   -- ++(right:1.8) -- ++(1.25em, 2.5em) -- ++(left:1.8) -- cycle;
    
    \draw [decorate, decoration={brace, amplitude=1ex, mirror}, very thick] (2.35, 0) -- ++(0, 3.5em);
    
    \node [img box, fill=black!20] at (3.75, 1.75em) (img 1) {};
    \node [img box, fill=black!20] at (6.3, 1.75em+-1.1cm) (img 1b) {};
    \node [font=\small] at (img 1) {image 1};
  \end{scope}
  
  \node [align=center, font=\small] at ($ (img 1)!0.5!(img 2) + (left:1.5) $)  {convert to\\greyscale};
  \node [align=center, font=\small] at ($ (img 1)!0.5!(img 2) + (right:0.8) $) {stack\\channels};
  \draw [very thick, -stealth] (img 1) -| (img 1b);
  \draw [very thick, -stealth] (img 2) -| (img 2b);
\end{tikzpicture}
\caption{Merging two images converted to greyscale by stacking them resulting in a two-channel image.\cite{Mohr21}} \label{fig:merging2:howto}
\end{center}
\end{figure}

The other way is to stack the images' channels. For an image with colour channels, this leads to a total of six channels. The process is illustrated in figure \ref{fig:merging2:howto}, where the images are converted to greyscale for better handling.

Both approaches lead to an altered tensor, that is delivered to the neural network. Therefore, attention must be paid to the distinctions between the input layer and images.
The influence of these two approaches is evaluated. With the same database and the same convolutional neural network the stacking approach of merging leads to a prediction accuracy of 98.36\%. When joining the images, as in figure \ref{fig:merging1:howto}, they only achieve an accuracy of 61.48\%. Because of these results, the approach with joined images will not be investigated any further.
Therefore, better results can be expected from a database with stacked images. When the colour channels of the photos are stacked, the important characteristics are in the same relative location. The performance of the CNN may be impacted even if a CNN can achieve a small invariance to translations through parameter sharing. In addition to this, the data are mixed up with the merged images because of convolutions and pooling. With stacked pictures, this is not the case since each operation is applied to each channel independently.

\subsection{Siamese Networks}

Siamese Networks (SNs) are neural networks that learn relationships between encoded representations of instance pairs that lie on the low dimensional manifold, where a chosen distance function $d\omega$ is used to find the similarity in output space. Below we briefly describe state-of-the-art convolutional SNs which have mainly been used for face verification and face recognition.

Sun et al. \cite{Sun14} presented a joint identification-verification approach for learning face verification with a contrastive loss and face recognition using cross-entropy loss. To balance loss signals for both identification and verification, they investigate the effects of varying weights controlled by $\lambda$ on the intra-personal and inter-personal variations.

\begin{figure}
\includegraphics[width=\textwidth]{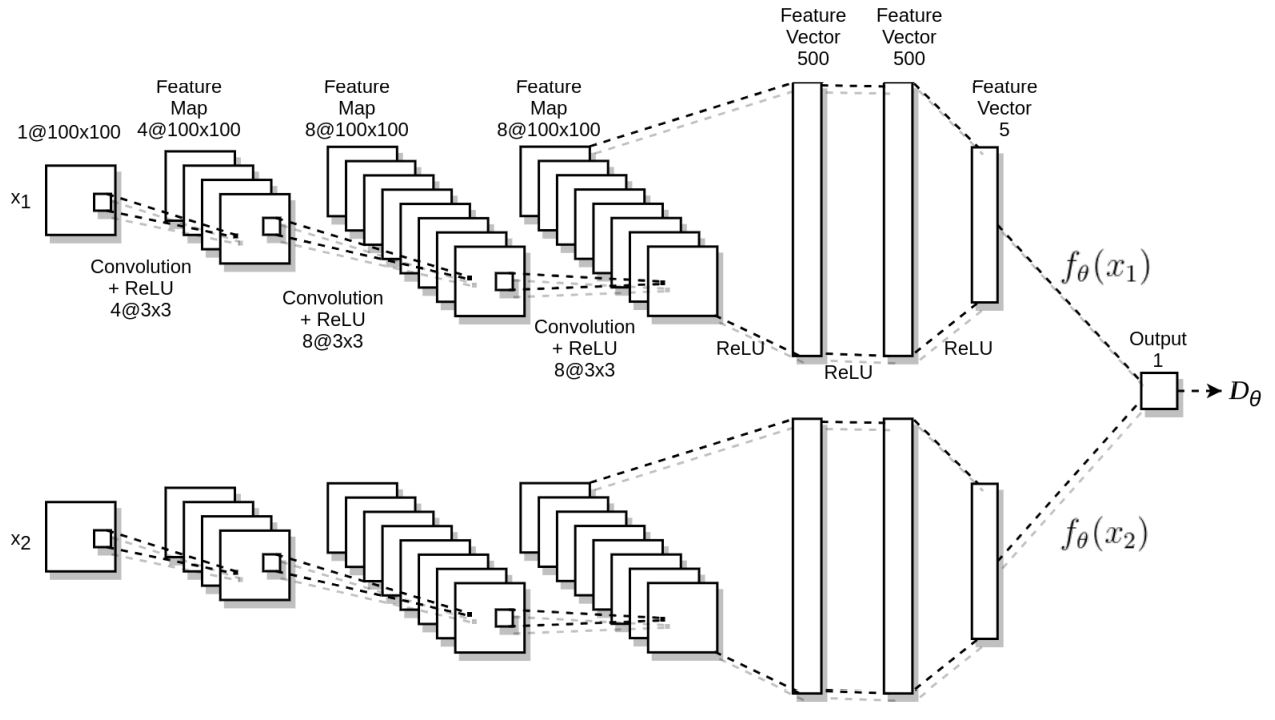}
\caption{Architecture of the Siamese Network which is used as a baseline\cite{Deshpande.2020}} \label{fig:SiameseNetwork}
\end{figure}

Wen et al. \cite{Wen16} propose a center loss function to improve discriminative feature learning. The center loss function proposed aims to improve the discriminability between feature representations by minimising the intra-class variation while keeping features from different classes separable. 

The center loss is given as 
\begin{equation}
    \mathcal{L} = - \sum^m_{i=1} \log(e^z) / (\sum^n_{j=1}e^z) + \lambda_2 \sum_m i=1||x_i-c_{y_i}||^2_2
\end{equation}

where $z = W^T_j x_i+b_j$.
The $c_{y_i}$ is the centroid of feature representations pertaining to the $i_{th}$ class.
This penalises the distance between class centers and minimises the intra-class variation while the softmax keeps the inter-class features separable. The centroids are computed during stochastic
gradient descent as full batch updates would not be feasible for large networks.

Liu et al. \cite{Liu17} proposed Sphereface, a hypersphere embedding that uses an angular softmax loss
that constrains discrimination on a hypersphere manifold, motivated by the prior that faces lie on a manifold. The model achieves competitive results on several datasets. 

The most relevant and notable use of Siamese Networks for face verification is the DeepFace
network, introduced by Taigman et al. \cite{Taigman.2014}. The performance obtained was on par with human-level performance on the Faces in the Wild (LFW) dataset and significantly outperformed previous
methods. However, it is worth noting this model is trained on a very large dataset from Facebook (SFC), therefore the model can be considered to be performing transfer learning before evaluation. The model also carries out some manual steps for detecting, aligning, and cropping faces from the images.
For detecting and aligning the face a 3D model is used. The images are normalized to avoid any
differences in illumination values, before creating a 3D model which is created by first identifying 6 fiducial points in the image using a Support Vector Regressor from an LBP histogram image descriptor.
Once the faces are cropped based on these points, a further 67 fiducial points are identified for a 3D mesh model, followed by a piece-wise affine transformation for each section of the image. The cropped image is then passed to 3 CNN layers with an initial max-pooling layer followed by two fully-connected layers. Similar to Capsule Networks, the authors refrain from using max pooling at each layer due to information loss.

The above work all achieve comparable state-of-the-art results for face verification using either a single CNN or a combination of various CNNs, some of which are pretrained on large related datasets.
A siamese network as proposed by Deshpande et al. \cite{Deshpande.2020} is used as a baseline to compare the results of the other approaches discussed in this paper. The architecture can be seen in Figure~\ref{fig:SiameseNetwork}. The two networks used are identical and the input to the model is a single channel or grayscale image pairs $x_1$ and $x_2$. Each module has three convolutional layers with a number of feature maps as 4, 8, and 8 from left to right respectively of size 100 × 100 each. The convolutional layers are followed by three fully connected layers of size 500, 500, and 5 respectively. The kernel size of 3 × 3 is used for convolutions with a stride of 1. The ReLU activation function is used on the output feature maps from each layer. During training, the contrastive loss function is used. The loss function is parameterised by the weights of the neural network $\theta$ and the training sample i. The $i_{th}$ training sample from the dataset is a tuple ($x_1$, $x_2$, y) where $x_1$ and $x_2$ are a pair of images and the label $y$ is equal to 1 if $x_1$ and $x_2$ belong to same class and 0 otherwise. The following equation describes the contrastive loss function
\begin{equation}
\label{eq:clf}
    L(\theta,(x_1,x_2,y)^i) = y \frac{1}{2} D^2_{\theta,i}+(1-y)\frac{1}{2}(\max \{ 0,m-D_{\theta,i} \})^2
\end{equation}
with $D_{\theta,1}=||f_\theta(x_1)-f_\theta(x_2)||_{2,i}$. The first term of the right hand side of equation~\ref{eq:clf} imposes cost on the network if the input image pair $x_1$ and $x_2$
belongs to the same class ($y=1$). The second term penalises the input sample if the data belongs to different classes ($y=0$).
A more in-depth explanation of the contrastive loss function can be found at \cite{Hadsell.2006}.

In contrast, this work looks to use a smaller Capsule Network that is more efficient, and requires little
preprocessing steps (no aligning, cropping, etc.) and can learn from less data, which is particularly evident in the fact that it achieves high accuracy on small data sets that are challenging for classical CNNs as shown in Section~\ref{sec:Experiments}.

\subsection{Siamese Network with Capsules}
\label{sec:SiameseCaps}

One model architecture, which seems especially suitable for one-shot identification tasks, is the capsule network architecture proposed by \cite{DynamicRouting}. Their potential comes from vectors, referred to as capsules, which contain properties that are intelligible to humans. In a specific approach called Dynamic Routing, the information transfer between the capsules is amplified or mitigated. During training, the vector output of the last capsules is masked and then inserted into a decoder to restore the perceived features of the input image which can be used to generate images.

\begin{figure*}
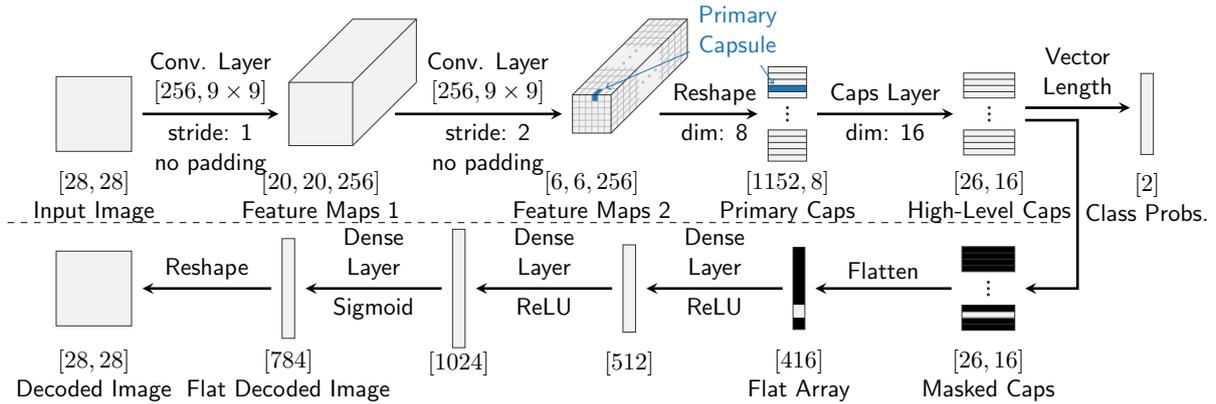

    \centering
    \includestandalone[width=\linewidth]{Figures/caps_net}
    \caption{Architecture of the used CapsNet.\cite{Quetscher.2022}}
    \label{fig:CapsNetArch}
\end{figure*}

Subsequently, we use the term \textit{CapsNet} for the model that consists of the capsule network itself and also the decoder.
The architecture of our CapsNet is illustrated in Figure~\ref{fig:CapsNetArch}. It is based on the architecture proposed by \cite{Quetscher.2022} with slight alterations. We use a Leaky-ReLU activation function with a leak of $a = 0.01$  is used for the two subsequent convolutional layers.

Each primary capsule consists of a group of $n_p$ feature maps. The number of values inside a capsule $n_p$ must be a divisor of the number of feature maps $n_m$ (here: $n_m = 256$), such that $\frac{n_m}{n_p}$ is the number of primary capsules per location. Together with the dimensions $[h_m, w_m, n_m]$ of the feature maps it results in the number $N_p$ of primary capsules \cite{Quetscher.2022}. 

\begin{equation}
    N_p = h_m \cdot w_m \cdot \frac{n_m}{n_p}
\end{equation}

Capsules also have an activation function, similar to the more commonly used neurons.
As in \cite{DynamicRouting}, we use the squashing function

\begin{equation}
    \hat{\mathbf g} = \frac{||\mathbf g||_2^2}{(1 + ||\mathbf g||_2^2)}, \frac{\mathbf g}{||\mathbf g||_2}
    \label{eq:Squashingfunction}
\end{equation}

where $\mathbf g$ is the vector of a capsule. It squashes the length of the output vector $\hat{\mathbf g}$ of a capsule between $0$ and $1$. The resulting length depends non-linearly on $\mathbf g$. The squashing function is performed on both primary and subsequent high-level capsules.

Since each primary capsule $i$ and each high-level capsule $j$ are entirely linked to one another by a weight matrix $\mathbf W_{ij}$, the values in a capsule may be thought of as neurons. There isn't a bias vector, though.
The Dynamic Routing algorithm \cite{DynamicRouting} is executed between the primary capsules and the high-level capsules. An additional coupling coefficient $c_{ij}$ is added between each primary capsule $i$ and each high-level capsule $j$, which stems from a routing logit $b_{ij}$ by applying a softmax across $j$. The routing logits $b_{ij}$ are initialised with zeros for each forward pass and updated within the routing iterations. Their values result from the relevance of the prediction of a primary capsule $i$ to the prediction of a high-level capsule $j$. As a result, the connection between both capsules is amplified or mitigated.

In this paper, a CapsNet as described is used as part of a siamese network as proposed by \cite{SiameseCapsule}. Two identical CapsNet´s are used in a setup like a siamese network and a contrastive loss function is used during training. This is a straightforward implementation of the original idea of siamese networks with mainly used CNNs replaced by capsule networks.

\section{Experimental Results}
\label{sec:Experiments}

In the following sections, experiments are conducted on three different data sets. A dataset with images from an industrial application, the smallNORB dataset, and the AT\&T database of faces. This provides a large variety in the experiments and tests the approaches in different application domains.

In all three of these experiments, three alternative strategies have been applied. A Siamese network is used as a baseline. A convolutional neural network, which was specially designed for stacked images, and additionally a Siamese network, which is based on capsule networks instead of convolutional neural networks. 10-fold cross-validation was used for the presented accuracies in every experiment except the industrial application.

\subsection{Industrial Application}

Anodes are produced in an automatic process in aluminum plants. Due to the need for high temperature and force, this process is expensive and time-consuming. One of the production steps during manufacturing an anode is to burn them in a special way in a kiln. During the burning, the anodes lie in irregular and not pre-defined ways on a conveyor belt. For reasons like traceability of defects, quality control, predictive maintenance, and quality assurance it is of high importance for the aluminum plant to be able to identify the anodes between initial jogging and the final electrolysis.

Anodes do undergo some surface transformation during baking. Additionally, burn-off during this procedure has the potential for harm to some locations. Moreover, the substance used to stabilise the anodes during baking might adhere to the surface of the anodes. In both instances, just a portion of the anode is influenced. Other places' features and surface textures are unaffected. In the event that tags are utilised, it is possible for one of the aforementioned processes to destroy the attached tag, rendering the tag useless for object identification and leaving this particular anode unrecognisable.

There is only one instance the anodes go through the manufacturing process, thus there isn't enough data produced to train a neural network to recognise each anode (every anode is a class for itself). Due to just having one picture of each anode accessible for training, a large number of anodes would need to be learned. Because of this, a neural network is to be trained to compare two anodes and make an educated guess, if both anodes are the same or not. This allows to create a neural network, that does not need to be retrained for every anode, thus solving the problem of not having enough training data. Only two images of every anode are required to identify an object.
For more detailed information on the dataset kindly refer to Mohr et. al. \cite{Mohr21}.

\begin{figure}
\begin{center}
\input{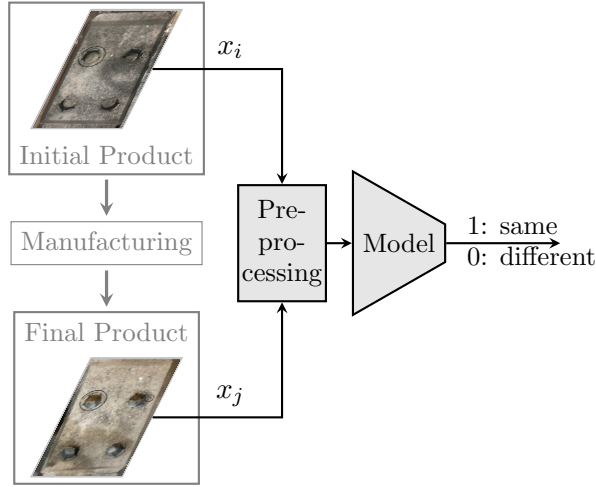} 
\caption{The workflow for the industrial application. During the manufacturing step several operations are done which alter the appearance of the anodes.} \label{fig:workflow}
\end{center}
\end{figure}

Every anode produced on the conveyor is captured using photographs. A novel picture is created and sent into the neural network each time an anode moves through the furnace. The neural network determines how certain it is that the two anodes presented to the network are identical by comparing this to every other anode. To compare the processed anode to all previously observed anodes, the network must be run through several times because it can only compare two pictures at a time. The two anodes with the highest likelihood are determined after all anodes have been observed. Unlike applications like facial recognition, only a small number of anodes cross the cameras at once. Figure \ref{fig:workflow} explains the whole process flow.

\subsubsection{Preprocessing and Data Augmentation}
\label{imageGeneration}

\begin{figure}
\begin{center}
\includegraphics[scale=0.5]{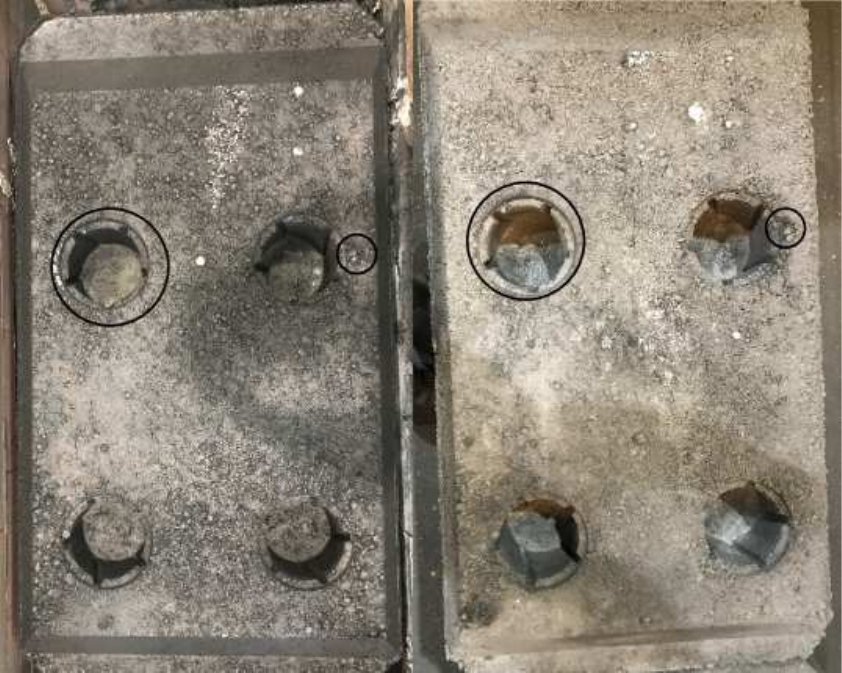}
\caption{A random example from the dataset showing variances in surface texture before (left) and after (right) the anode is baked in the furnace.\cite{Mohr21}} \label{fig:AnodeBeforeAndAfter}
\end{center}
\end{figure}

During ongoing production, a number of images of anodes were taken. The images were taken before and after the baking in the kiln and various other manufacturing steps. In Figure~\ref{fig:AnodeBeforeAndAfter} an example of these images can be seen. The number of images is limited because the manufacturing of one anode takes a significant amount of time. Therefore images are generated and the changes in the appearance of the anodes are simulated to enhance the dataset and increase its size.
 
As can be seen in Figure \ref{fig:AnodeBeforeAndAfter} significant features on the surface of the anode remain after processing. Reviewing the available pictures of anodes before and after the baking process allowed to analyse and identify the most common changes and similarities in an anode during production.

During the green anode's production, the stubs are formed. Because the device used to make the stub is removed after the anode is formed, the alignment of these always varies between two anodes. Between every processing operation, this characteristic is maintained. During the furnace's baking process, the surface's brightness and texture change. The anodes are also added to and altered with spots, markings, and other modifications during processing.

\begin{figure}
\begin{center}
\includegraphics[scale=0.37]{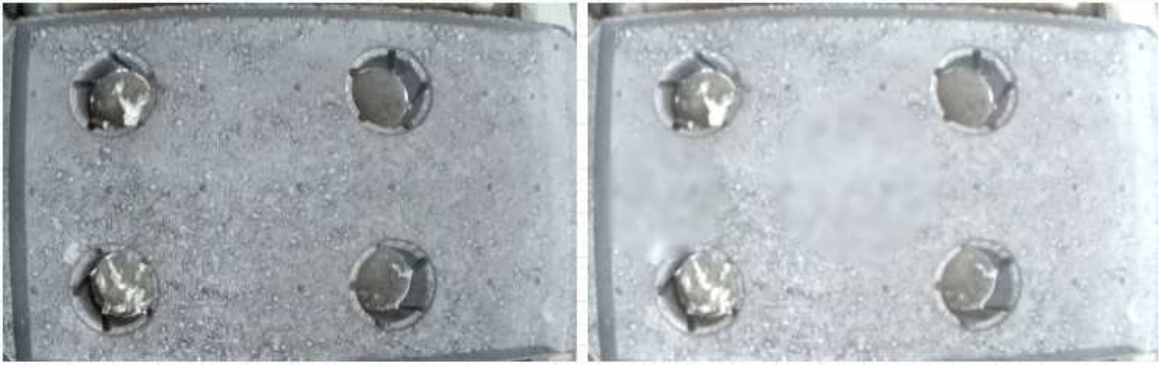}
\caption{Generated image as it is used in the training data set. This image was generated out of a photo of an anode with data augmentation techniques.\cite{Mohr21}} \label{fig:AnodeSimulated}
\end{center}
\end{figure}

To overcome the problem of limited quantity and limited diversity of data the existing images are manipulated with affine transformations. For simulating the variances all images of unaltered anodes are manipulated. Each image in the data set is rotated randomly about its centre. Brightness and fine contours are changed. Count and position of burn-offs and build-ups are manipulated randomly and simulated with blurred circles, that are layered over the anodes. These images are saved separately and used as part of the training data set. The data set is combined out of simulated and original images. Figure \ref{fig:AnodeSimulated} exemplarily shows a simulated image of an anode next to the original image.
 
\subsubsection{Results}

\begin{figure*}
\includegraphics[width=\textwidth]{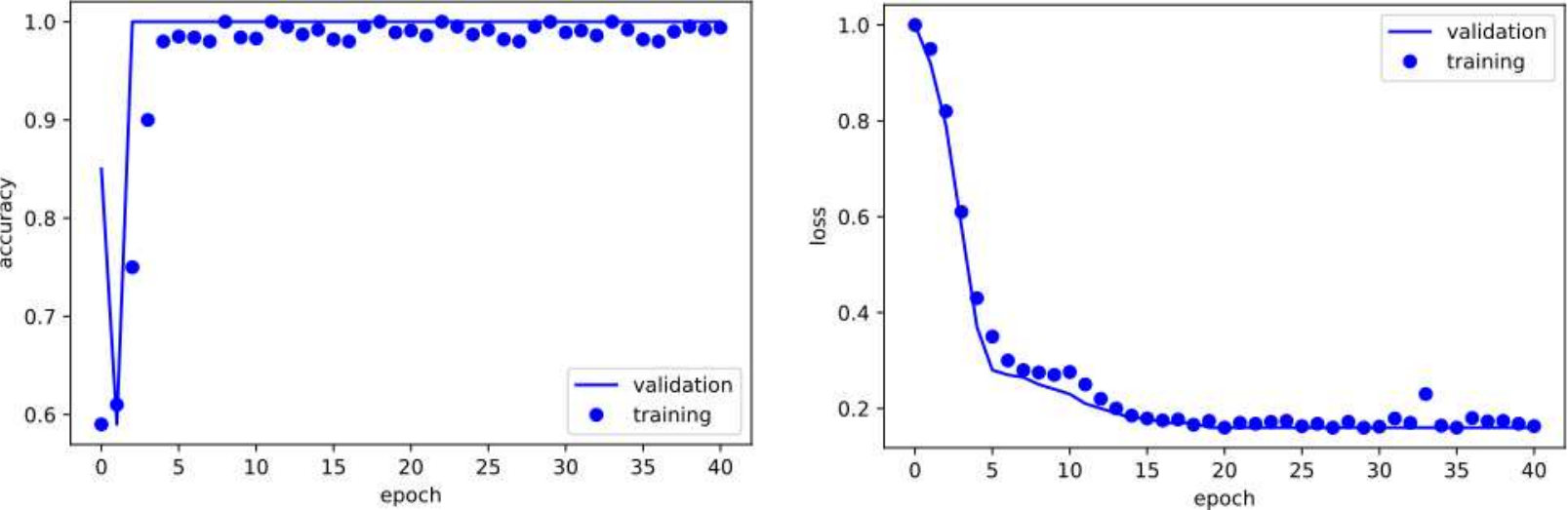}
\caption{The accuracy (left) and the loss (right) during training for 40 epochs. The shown development stems from training the approach for merged images on the industrial application dataset.} \label{fig:LearningBehaviour}
\end{figure*}

Our experiments have shown, that stacking images achieves high accuracy. Further experiments have displayed, that the convolutional neural network reaches an accuracy of 98.4\% when the early stopping terminated the training process. As \ref{fig:LearningBehaviour} illustrates this high level is reached in a very stable way. No signs of overfitting can be found as new images are also identified correctly, and the accuracy of the validation data does oscillate strongly or even shrink during the training. 
Figure \ref{fig:LearningBehaviour} shows the learning behaviour based on the accuracy and loss during training.

The siamese network was trained on the same training data and the accuracy was achieved on the same set of test images as used for training the neural network with the merged images. It reaches an accuracy of 96.4\%.

The approach with merged images for one-shot identification has the main advantage over state-of-the-art siamese networks. Siamese networks have a more complex architecture regarding the prediction because the images are forwarded through two neural networks instead of just one. In industrial applications with a small batch of objects to identify just once -- in opposite to face recognition where siamese networks are very common -- pre-processing images through the network and storing outputs and then in an additional step calculating the similarity is less helpful. Therefore, in such industrial applications, they are often slower regarding prediction. Fast predictions are of the utmost importance in industrial applications because in general, they have some kind of real-time requirements.

The siamese CapsNet is tested in two different settings for the industrial application. At first, the same dataset is used which is used for the merged images and the siamese network. On this dataset, the siamese CapsNet achieves an accuracy of 96.4\%.
Additionally, we used the decoder network of the CapsNet which gives it the ability to generate images after a training phase\cite{Mohr.2021.capsules}. The CapsNet is trained on the original images of the anodes for 20 epochs. After this training, the decoder network is used to generate the same amount of images as previously generated manually. A new dataset is composed of the original and the generated images and the siamese CapsNet is trained with the same architecture and parameters achieving an accuracy of 98.5\%.

\subsection{Results on smallNORB Dataset}

\begin{figure}
\label{fig:smallNORB}
\includegraphics[width=\textwidth]{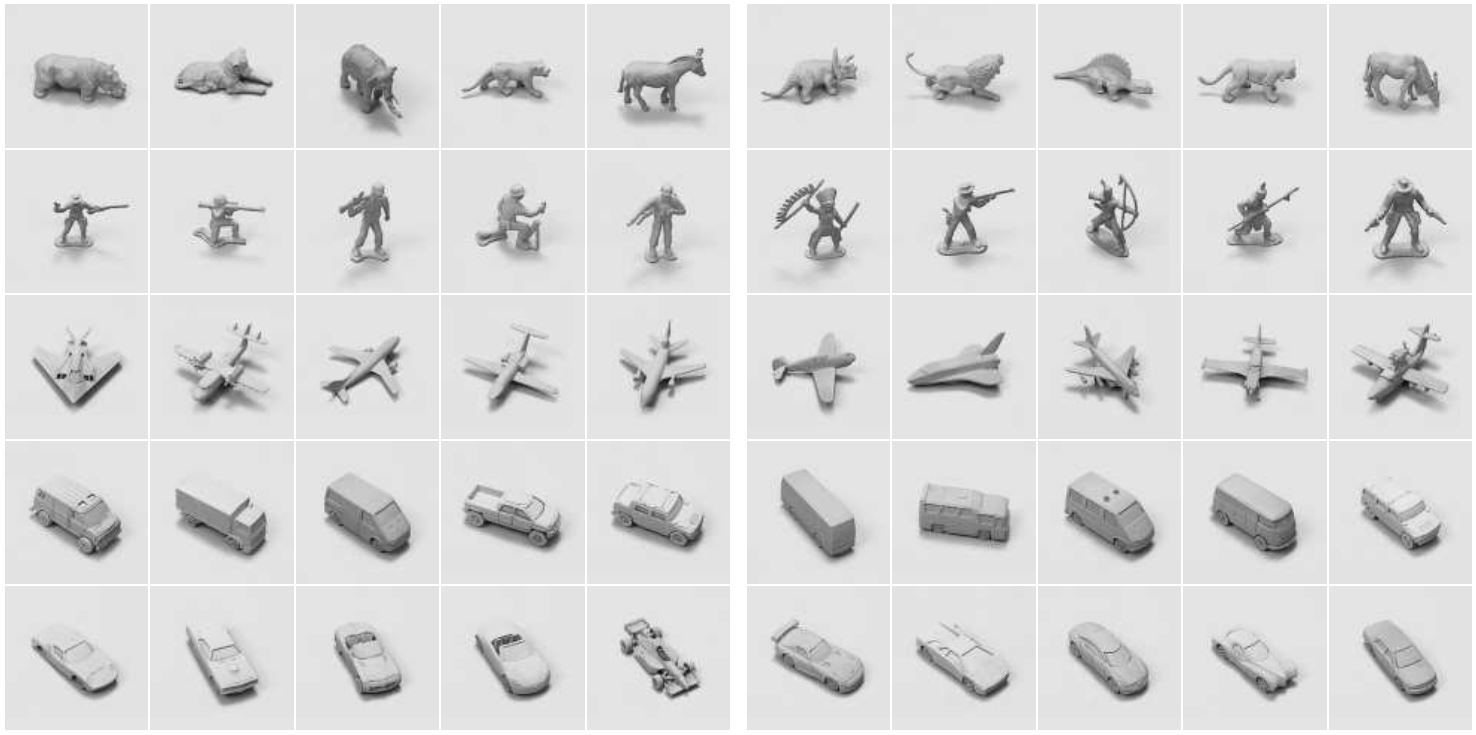}
\caption{Random examples from the smallNORB dataset with examples for all 5 categories from the training (left) and test set (right).\cite{LeCun.2004}}
\end{figure}

The dataset smallNORB \cite{LeCun.2004} is a collection of 48600 stereo, grayscale images (96 × 96 × 2), representing 50 plastic toys belonging to 5 generic categories: humans, airplanes, trucks, cars, and four-legged animals. Each toy was photographed by two cameras under 6 lighting conditions, 9 elevations, and 18 azimuths. A turntable was used to move the toys and they were painted in a uniform green colour. The dataset is split in half; 5 instances of each category for the training and the remaining ones for the testing leading to a few-shot learning task. Examples can be seen in Figure~\ref{fig:smallNORB}

\begin{figure*}
\begin{center}
\includegraphics[width=\textwidth]{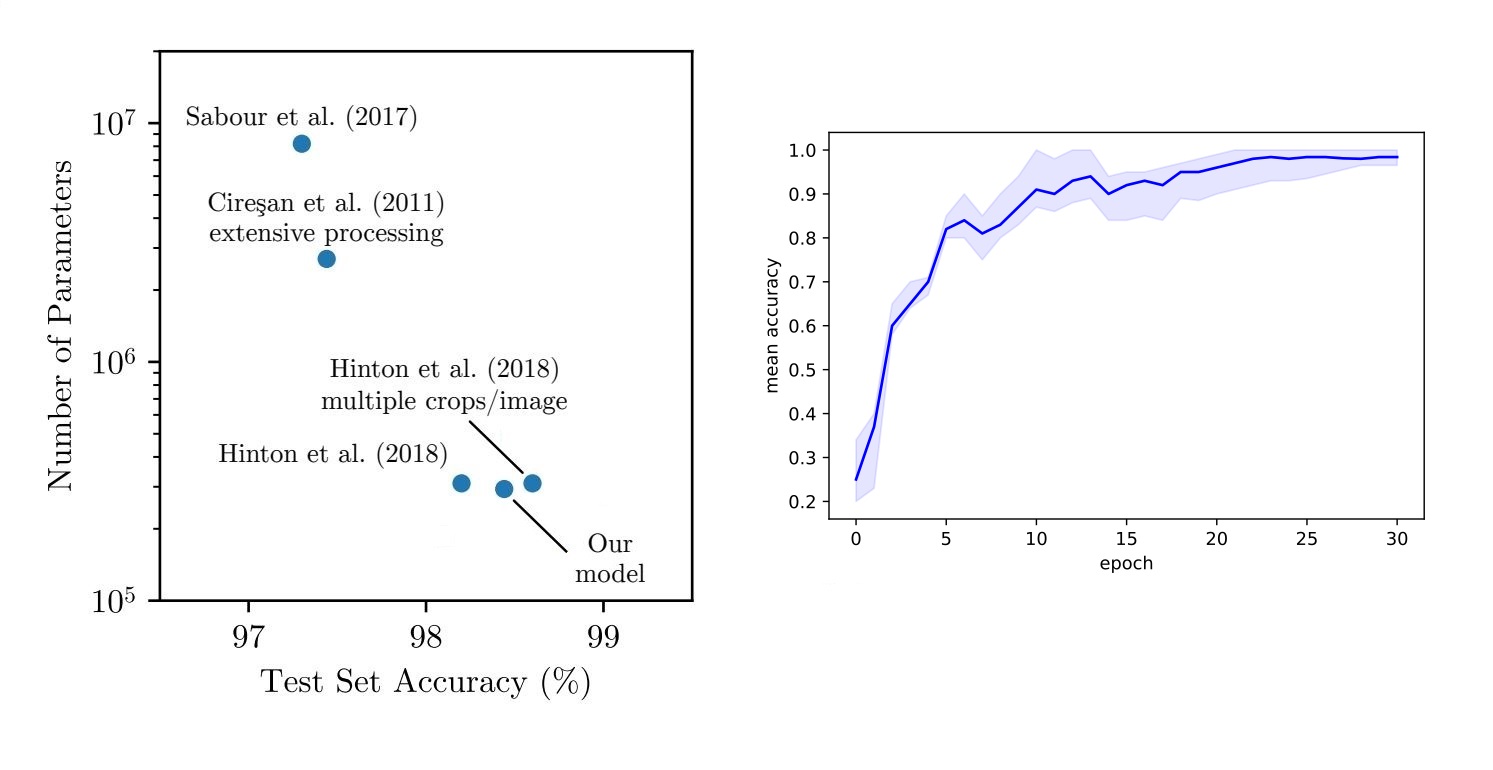}
\caption{Test set accuracy and number of parameters of models that have achieved state-of-the-art
results on smallNORB visual recognition on the left and development of the accuracy of the siamese capsule network on the right. SmallNORB does not have a validation split.}\label{fig:smallNORB_accuracy}
\end{center}
\end{figure*}

The baseline siamese networks reach an accuracy of 92.5\% and the approach with merged images of 94.7\%.
That the approach with merged images outperforms the baseline siamese network shows the strength and versatility of this approach as it can outperform our baseline. However, it is a different setting, which is different from the setting for which this approach was developed.
We used the same architecture as for the industrial application for the smallNORB dataset for the approach with siamese capsule networks. With this, we received an accuracy of 98.4\% translating to an error rate of only 1.6\% which is on par with the state-of-the-art \cite{Ciresan.2011} with 30 epochs. Out of the used approaches, siamese capsule networks perform best on this dataset which shows how well capsule networks can perform even in complex tasks like face recognition and zero-shot learning tasks. Refer to Figure~\ref{fig:smallNORB_accuracy} for visualisation and comparison to other approaches. In the comparison, a selection of other approaches to the smallNORB dataset can be seen which are sorted based on accuracy and number of parameters. Some of these other approaches have extensively preprocessed or cropped the images.

\subsection{Results on AT\&T Database of Faces}
 The AT$\&$T face recognition and verification dataset consists of 40 different
subjects with only 10 grey-pixel images per subject in a controlled setting. The pictures have been taken between April 1992 and April 1994. For some subjects, the images were taken at different times, varying the lighting, facial expressions and additional features like glasses. All the images were taken against a dark homogeneous background with the subjects in an upright, frontal position. Examples can be seen in Figure~\ref{fig:approach}.

\begin{figure}
\centering
\includegraphics[scale=0.8]{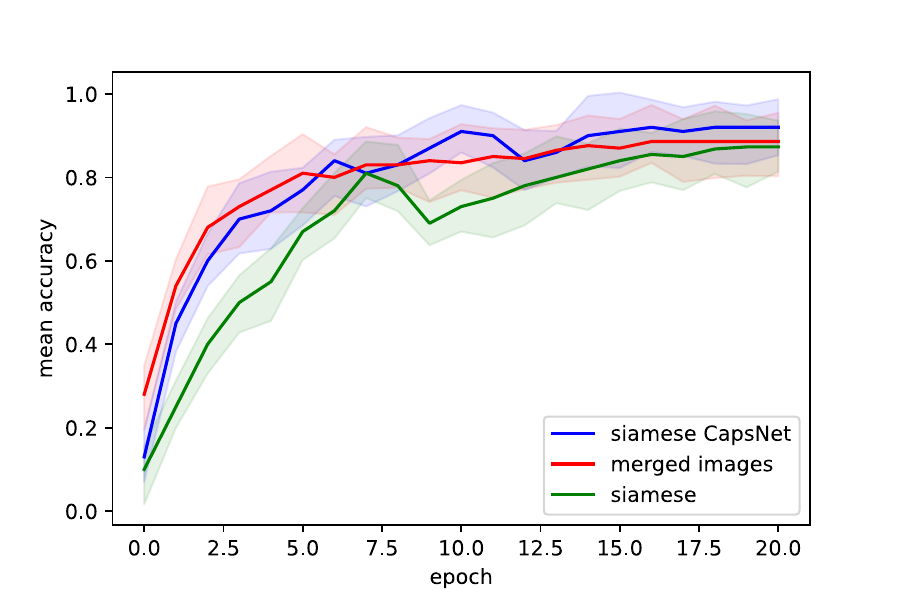}
\caption{Mean accuracy after each epoch of training for all three approaches. Shaded area denotes deviation.}
\label{fig:faces_accuracy}
\end{figure}

This smaller dataset allows us to test how the approaches perform with little data. For testing, we hold out 5 subjects so that we are testing on unseen subjects, as opposed to training on a given viewpoint of a subject and testing on another viewpoint of the same subject. Hence, zero-shot pairwise prediction is performed during testing.
The siamese network achieves an accuracy of 87.3\% and the CNN with merged images 88.6\%, which shows that this approach would require more tuning and adaptation because the face recognition task highly differs from the industrial application and the approach was specifically designed for the requirements of this application without paying attention to generalisation or performance on other tasks.
As expected the siamese CapsNet achieves the best accuracy with 90.2\% showing as already expected and explained in Section~\ref{sec:SiameseCaps} that CapsNet´s perform very well with little data making them suitable for particular tasks and settings like tiny data. Figure~\ref{fig:faces_accuracy} shows the development of the accuracy for all three approaches on this dataset.

\begin{table}
\caption{Results for the experiments and approaches in this paper presented as accuracy in percent}
\begin{center}
\begin{tabular}{|l|c|c|c|l|}
\hline
Approach & Industrial Dataset & \mbox{ smallNORB } & \mbox{ AT\&T faces }\\
\hline
merged images & 98.4\% & 94.7\% & 88.6\% \\
siamese & 96.4\% & 92.5\% & 87.3\% \\
siamese CapsNet & 97.9\% & 98.4\% & 90.2\% \\
\hline
\end{tabular}
\end{center}
\end{table}

\section{Conclusion and Future Prospects}
In summary, we have presented three different approaches to dealing with limited data and one-shot or zero-shot tasks. These approaches were tested on data sets of different sizes and from different domains including an industrial application and a face recognition task.
Overall, the approach with the siamese capsule networks achieves the best results in terms of accuracy. This approach is only narrowly beaten by the approach with merged images in the industrial application specifically designed for it.

All experiments in this paper show that the amount of available data has a significant influence on the achievable accuracy and that special approaches have a right to exist for special tasks. It can also be seen that there are some promising approaches for dealing with small data sets and one-shot tasks. Here, the performance of the recently proposed capsule networks is particularly noteworthy. As these are still in the early stages of development, it is expected that the achievable accuracy and number of parameters will improve further with future work.

These capsule networks also provide ideas for further development in the context of one-shot tasks. It has already been shown that the decoder network can be used for continuous learning when training on large data sets. In addition, due to the high comparability of the capsules with elements of human vision, high explanatory power can be achieved. It is valuable to investigate how this explanatory power can also be applied to one-shot and especially face recognition tasks.

\subsection{Acknowledgements} This work was funded by the federal state of North Rhine-Westphalia and the European Regional Development Fund FKZ: EFRE-040021. The industrial application was supported by Trimet Aluminium SE.

\bibliographystyle{unsrt}
\bibliography{bibliography}

@InProceedings{LeCun1998,
  author    = {Yann LeCun and Leon Bottou and Yoshua Bengio and Patrick Haffner},
  title     = {Gradient-Based Learning Applied to Document Recognition},
  booktitle = {Proceedings of the IEEE},
  year      = {1998},
}

@phdthesis{Tielenman.2014,
 author = {Tielenman, Tijmen},
 year = {2014},
 title = {{Optimizing Neural Networks That Generate Images}},
 publisher = {Department of Computer Science},
 school = {{University of Toronto}},
 type = {{Dissertation}}
}

@Article{Brusey2009,
  author  = {James Brusey and Duncan C Mcfarlane},
  title   = {Effective RFID-based Object Tracking for Manufacturing},
  journal = {International Journal of Computer Integrated Manufacturing},
  year    = {2009},
  pages   = {638-647},
}

@Article{Benhimane2008,
  author  = {Selim Benhimane and Hesam Najafi and Matthias Grundmann and Ezio Malis and Yakup Genc and Nassir Navab},
  title   = {Real-Time Object Detection and Tracking For Industrial Applications},
  journal = {VISAPP 2008: Third International Conference on Compter Vision Theory and Applications},
  year    = {2008},
  volume  = {2},
  month   = {01},
  }

@Article{Minderer2019,
  author  = {Matthias Minderer and Chen Sun and Ruben Villegas and Forrester Cole and Kevin Murphy and Honglak Lee},
  title   = {Unsupervised Learning of Object Structure and Dynamics from Videos},
  journal = {33rd Conference on Neural Information Processing Systems},
  year    = {2019},
}

@article{Mahieu2019,
  title={Carbon Block Tracking Package Based on Vision Technology},
  author={Pierre Mahieu and Xavier Genin and Christophe Bouch{\'e} and David Brismalein and Herv{\'e} P{\'e}droli},
  journal={Light Metals},
  year={2019}
}

@article{Koch.2015,
 author = {Koch, Gregory and Zemel, Richard and Salakhutdinov, Ruslan},
 year = {2015},
 title = {Siamese Neural Networks for One-shot Image Recognition},
 journal = {32nd International Conference on Machine Learning}
}

@article{Chicco.2020,
 author = {Chicco, Davide},
 year = {2020},
 title = {{Siamese Neural Networks: An Overview}},
 journal = {{Methods in molecular biology}}
}

@article{Deshpande.2020,
title = {One-Shot Recognition of Manufacturing Defects in Steel Surfaces},
journal = {48th SME North American Manufacturing Research Conference, NAMRC 48},
year = {2020},
author = {Aditya M. Deshpande and Ali A. Minai and Manish Kumar}
}

@article{Kuhl.2020,
  author    = {Niklas K{\"{u}}hl and
               Marc Goutier and
               Lucas Baier and
               Clemens Wolff and
               Dominik Martin},
  title     = {Human vs. supervised machine learning: Who learns patterns faster?},
  journal = {ArXiv},
  year      = {2020}
}

@article{Scherer.2010,
author = {Scherer, Dominik and Müller, Andreas and Behnke, Sven},
year = {2010},
month = {01},
pages = {92-101},
journal = {20th International Conference on Artificial Neural Networks (ICANN)},
title = {Evaluation of Pooling Operations in Convolutional Architectures for Object Recognition}
}

@article{Graves.2013,
  author    = {Alex Graves},
  title     = {Generating Sequences With Recurrent Neural Networks},
  year      = {2013},
  journal = {ArXiv}
}

@article{Cha.2017,
author = {Cha, Young-Jin and Choi, Wooram and Buyukozturk, Oral},
year = {2017},
month = {03},
pages = {361-378},
title = {Deep Learning-Based Crack Damage Detection Using Convolutional Neural Networks},
volume = {32},
journal = {Computer-Aided Civil and Infrastructure Engineering},
}

@article{Abdel.2003,
author = {Abdel-Qader, Ikhlas and Abudayyeh, Osama and Kelly, Michael},
year = {2003},
month = {10},
title = {Analysis of Edge-Detection Techniques for Crack Identification in Bridges},
journal = {Journal of Computing in Civil Engineering}
}

@inproceedings{Bromley.1993,
author = {Bromley, Jane and Guyon, Isabelle and LeCun, Yann and S\"{a}ckinger, Eduard and Shah, Roopak},
title = {Signature Verification Using a "Siamese" Time Delay Neural Network},
year = {1993},
booktitle = {Proceedings of the 6th International Conference on Neural Information Processing Systems},
series = {NIPS'93}
}

@inproceedings{Chopra.2005,
author = {Chopra, Sumit and Hadsell, Raia and Lecun, Yann},
year = {2005},
month = {07},
title = {Learning a similarity metric discriminatively, with application to face verification},
booktitle = {Proc. Computer Vision and Pattern Recognition}
}

@inproceedings{Taigman.2014,
author = {Taigman, Yaniv and Yang, Ming and Ranzato, Marc'Aurelio and Wolf, Lior},
year = {2014},
month = {09},
title = {DeepFace: Closing the Gap to Human-Level Performance in Face Verification},
booktitle = {Proceedings of the IEEE Computer Society Conference on Computer Vision and Pattern Recognition}
}

@inproceedings{DynamicRouting,
    author = {Sabour, Sara and Frosst, Nicholas and Hinton, Geoffrey E.},
    title = {Dynamic Routing between Capsules},
    year = {2017},
    booktitle = {Proceedings of the 31st International Conference on Neural Information Processing Systems},
    series = {NIPS'17}
    }

@misc{SiameseCapsule,
  author = {Neill, James O'},
  title = {Siamese Capsule Networks},
  year = {2018}
}

@article{Sun14,
  author    = {Yi Sun and
               Xiaogang Wang and
               Xiaoou Tang},
  title     = {Deep Learning Face Representation by Joint Identification-Verification},
  journal = {Neural Information Processing NIPS},
  year      = {2014}
}

@article{Wen16,
  title={A Discriminative Feature Learning Approach for Deep Face Recognition},
  author={Yandong Wen and Kaipeng Zhang and Zhifeng Li and Yu Qiao},
  journal={ECCV},
  year={2016}
}

@article{Liu17,
  author    = {Weiyang Liu and
               Yandong Wen and
               Zhiding Yu and
               Ming Li and
               Bhiksha Raj and
               Le Song},
  title     = {SphereFace: Deep Hypersphere Embedding for Face Recognition},
  journal = {IEEE Conference on Computer Vision and Pattern Recognition (CVPR)},
  year      = {2017}
}

@article{Mohr21,
author = {Mohr, Janis and Breidenbach, Finn and Frochte, Jörg},
year = {2021},
month = {10},
title = {An Approach to One-Shot Identification With Neural Networks},
journal={IJCCI}
}

@article{abeysinghe2021capsule,
  title={Capsule Networks for Character Recognition in Low Resource Languages},
  author={Abeysinghe, C and Perera, I and Meedeniya, DA},
  journal={Machine Vision Inspection Systems, Volume 2: Machine Learning-Based Approaches},
  pages={23--46},
  year={2021},
  publisher={Wiley Online Library}
}

@article{gagana2018activation,
  title={Activation function optimizations for capsule networks},
  author={Gagana, B and Athri, HA Ujjwal and Natarajan, S},
  journal={2018 International Conference on Advances in Computing, Communications and Informatics (ICACCI)},
  pages={1172--1178},
  year={2018},
  organization={IEEE}
}

@article{GeoffreyE.Hinton.1981,
 author = {{Geoffrey E. Hinton}},
 year = {1981},
 title = {{A Parallel Computation that Assigns Canonical Object-Based Frames of Reference}},
 journal = {Seventh International Joint Conference on Artificial Intelligence}
}

@article{GeoffreyE.Hinton.1981b,
 author = {{Geoffrey E. Hinton}},
 year = {1981},
 title = {{Shape Representation in Parallel Systems}},
 journal = {Seventh International Joint Conference on Artificial Intelligence}
}

@article{GeoffreyE.Hinton.2011,
 author = {{Geoffrey E. Hinton} and {A. Krizhevsky} and {S. Wang}},
 year = {2011},
 title = {{Transforming Auto-Encoders}},
 journal = {International Conference on Artificial Neural Networks}
}

@misc{Iesmantas.2018,
 author = {Iesmantas, Tomas and Alzbutas, Robertas},
 year = {2018},
 title = {{Convolutional capsule network for classification of breast cancer histology images}}
}

@misc{Mobiny.2018,
 author = {Mobiny, Aryan and {van Nguyen}, Hien},
 year = {2018},
 title = {{Fast CapsNet for Lung Cancer Screening}}
}

@article{Afshar.2018,
 author = {Afshar, Parnian and Mohammadi, Arash and Plataniotis, Konstantinos N.},
 year = {2018},
 title = {{Brain Tumor Type Classification via Capsule Networks}},
 journal = {25th IEEE International Conference on Image Processing ICIP}
}

@article{Kumar.2018,
 author = {Kumar, Amara Dinesh},
 title = {{Novel Deep Learning Model for Traffic Sign Detection Using Capsule Networks}},
 year = {2018},
 journal = {{International Journal of Pure and Applied Mathematics Volume 118 No. 20}}
}

@article{Renkens.2018,
 author = {Renkens, Vincent and {van hamme}, Hugo},
 year = {2018},
 title = {{Capsule Networks for Low Resource Spoken Language Understanding}},
 journal = {Proc. Interspeech 2018}
}

@article{Xi.2017,
 author = {Xi, Edgar and Bing, Selina and Jin, Yang},
 year = {2017},
 title = {{Capsule Network Performance on Complex Data}},
 journal = {International Joint Conference on Neural Networks (IJCNN)}
}

@article{Rajasegaran.2019,
 author = {Rajasegaran, Jathushan and Jayasundara, Vinoj and Jayasekara, Sandaru and Jayasekara, Hirunima and Seneviratne, Suranga and Rodrigo, Ranga},
 year = {2019},
 title = {{DeepCaps: Going Deeper with Capsule Networks}},
 journal = {IEEE/CVF Conference on Computer Vision and Pattern Recognition CVPR}
}

@article{Ciresan.2011,
  author    = {Dan C. Ciresan and
               Ueli Meier and
               Jonathan Masci and
               Luca Maria Gambardella and
               J{\"{u}}rgen Schmidhuber},
  title     = {High-Performance Neural Networks for Visual Object Classification},
  journal = {Technical Report No. IDSIA-01-11 },
  year      = {2011}
}

@article{LeCun.2004,
author = {Lecun, Yann and Huang, Fu and Bottou, L.},
year = {2004},
title = {Learning methods for generic object recognition with invariance to pose and lighting},
journal = {IEEE Conference on Computer Vision and Pattern Recognition, CVPR}
}

@article{Hadsell.2006,
  title={Dimensionality Reduction by Learning an Invariant Mapping},
  author={Raia Hadsell and Sumit Chopra and Yann LeCun},
  journal={2006 IEEE Computer Society Conference on Computer Vision and Pattern Recognition (CVPR'06)},
  year={2006},
}

@article{Quetscher.2022,
  author    = {Felizia Quetscher and
               Christoph Kaufmann and
               Jörg Frochte},
  title     = {Investigation of Capsule Networks Regarding their Potential of Explainability and Image Rankings},
  journal   = {ICAART},
  year      = {2022}
}

@misc{Renzulli.2022,
  author = {Renzulli, Riccardo and Tartaglione, Enzo and Grangetto, Marco},
  title = {REM: Routing Entropy Minimization for Capsule Networks},
  year = {2022}
}

@article{Atefeh.2018,
  author    = {Atefeh Shahroudnejad and
               Arash Mohammadi and
               Konstantinos N. Plataniotis},
  title     = {Improved Explainability of Capsule Networks: Relevance Path by Agreement},
  year      = {2018},
  journal = {IEEE Global Conference on Signal and Information Processing (GlobalSIP)}
}

@article{Li.2019,
  author    = {Chenliang Li and
               Cong Quan and
               Li Peng and
               Yunwei Qi and
               Yuming Deng and
               Libing Wu},
  title     = {A Capsule Network for Recommendation and Explaining What You Like
               and Dislike},
  journal = {Proceedings of the 42nd International ACM SIGIR Conference on Research and Development in Information Retrieval},
  year      = {2019}
}

@inproceedings{Mohr.2021.capsules,
author = {Mohr, Janis and Tousside, Basile and Schmidt, Marco and Frochte, Jörg},
year = {2021},
month = {10},
title = {Explainability and Continuous Learning with Capsule Networks},
booktitle = {Proceedings of the 13th International Joint Conference on Knowledge Discovery, Knowledge Engineering and Knowledge Management}
}

@article{Lake2014OneshotLO,
  title={One-shot learning of generative speech concepts},
  author={Brenden M. Lake and Chia-ying Lee and James R. Glass and Joshua B. Tenenbaum},
  journal={Cognitive Science},
  year={2014},
  volume={36}
}

@article{Wu2012OneSL,
  title={One shot learning gesture recognition from RGBD images},
  author={Di Wu and Fan Zhu and Ling Shao},
  journal={2012 IEEE Computer Society Conference on Computer Vision and Pattern Recognition Workshops},
  year={2012}
}

@inproceedings{Lake2013OneshotLB,
  title={One-shot learning by inverting a compositional causal process},
  author={Brenden M. Lake and Ruslan Salakhutdinov and Joshua B. Tenenbaum},
  booktitle={NIPS},
  year={2013}
}

@article{Lake2012ConceptLA,
  title={Concept learning as motor program induction: A large-scale empirical study},
  author={Brenden M. Lake and Ruslan Salakhutdinov and Joshua B. Tenenbaum},
  journal={Cognitive Science},
  year={2012},
  volume={34}
}

@article{Lake2011OneSL,
  title={One shot learning of simple visual concepts},
  author={Brenden M. Lake and Ruslan Salakhutdinov and Jason Gross and Joshua B. Tenenbaum},
  journal={Cognitive Science},
  year={2011},
  volume={33}
}

@inproceedings{FeiFei.2003,
author = {Fei-Fei, Li and Fergus, Rob and Perona, Pietro},
title = {A Bayesian Approach to Unsupervised One-Shot Learning of Object Categories},
year = {2003},
publisher = {IEEE Computer Society},
booktitle = {Proceedings of the Ninth IEEE International Conference on Computer Vision - Volume 2},
pages = {1134},
series = {ICCV '03}
}

@article{FeiFei.2006,
  title={One-shot learning of object categories},
  author={Li Fei-Fei and Rob Fergus and Pietro Perona},
  journal={IEEE Transactions on Pattern Analysis and Machine Intelligence},
  year={2006}
}

\end{document}